\documentclass[sigconf, nonacm, screen]{acmart}

\AtBeginDocument{%
  \providecommand\BibTeX{{%
    \normalfont B\kern-0.5em{\scshape i\kern-0.25em b}\kern-0.8em\TeX}}}

\usepackage{xcolor}
\usepackage{multirow}
\usepackage{array, graphicx}
\usepackage{float}
\usepackage{subfig, amsmath}
\usepackage{geometry}
\usepackage{fancyhdr}

\acmConference[COTB'20]{Computer on the Beach}{01--03 de Abril de 2020}{Balneário
Camboriú, SC, Brazil}

\begin{document}

% \startpage{1}

\title[Can Giraffes Become Birds?]{Can Giraffes Become Birds?\\An Evaluation of Image-to-image Translation\\for Data Generation}

\author{Daniel V. Ruiz}
\authornote{Both authors contributed equally to this research.}
\email{dvruiz@inf.ufpr.br}
\affiliation{%
  \institution{Federal University of Paran\'{a}}
  \city{Curitiba}
  \state{Paran\'{a}}
}
\author{Gabriel Salomon}
\authornotemark[1]
\email{gsaniceto@inf.ufpr.br}
\affiliation{%
  \institution{Federal University of Paran\'{a}}
  \city{Curitiba}
  \state{Paran\'{a}}
}

\author{Eduardo Todt}
\email{todt@inf.ufpr.br}
\affiliation{%
  \institution{Federal University of Paran\'{a}}
  \city{Curitiba}
  \state{Paran\'{a}}
}

\renewcommand{\shortauthors}{Ruiz and Salomon, et al.}

\begin{abstract}
There is an increasing interest in image-to-image translation with applications ranging from generating maps from satellite images to creating entire clothes' images from only contours. In the present work, we investigate image-to-image translation using Generative Adversarial Networks (GANs) for generating new data, taking as a case study the morphing of giraffes images into bird images. Morphing a giraffe into a bird is a challenging task, as they have different scales, textures, and morphology. An unsupervised cross-domain translator entitled InstaGAN was trained on giraffes and birds, along with their respective masks, to learn translation between both domains. A dataset of synthetic bird images was generated using translation from originally giraffe images while preserving the original spatial arrangement and background. It is important to stress that the generated birds do not exist, being only the result of a latent representation learned by InstaGAN. Two subsets of common literature datasets were used for training the GAN and generating the translated images: COCO and Caltech-UCSD Birds 200-2011. To evaluate the realness and quality of the generated images and masks, qualitative and quantitative analyses were made. For the quantitative analysis, a pre-trained Mask R-CNN was used for the detection and segmentation of birds on Pascal VOC, Caltech-UCSD Birds 200-2011, and our new dataset entitled FakeSet. The generated dataset achieved detection and segmentation results close to the real datasets, suggesting that the generated images are realistic enough to be detected and segmented by a state-of-the-art deep neural network.

\end{abstract}
% \startpage{1}

% \keywords{datasets, neural networks, gaze detection, text tagging}




\renewcommand{\headrulewidth}{0.5pt}
\fancyhead[L]{\textbf{XI Computer on the Beach} \\ \textit{April 1--3, 2020,
Baln. Cambori\'{u}, SC, Brazil}}
\fancyhead[R]{\shortauthors}
\fancyfoot[C]{\thepage}

\maketitle

\fancypagestyle{mystyle}{
    \fancyhf{} 
    \renewcommand{\headrulewidth}{0pt} 
    \renewcommand{\footrulewidth}{0pt} 
    \renewcommand{\footruleskip}{7mm} 
    \fancyfoot[L]{\footnotesize \textbf{Reference:}
    RUIZ, Daniel V., SALOMON, Gabriel, TODT, Eduardo. Can Giraffes Become Birds? An Evaluation of Image-to-image Translation for Data Generation. In:
    COMPUTER ON THE BEACH (COTB'20), 11., 2020, Balne\'{a}rio Cambori\'{u}. \textbf{Proceedings...}
    Balne\'{a}rio Cambori\'{u}: Vale do Itaja\'{i} University, 2020, p. 1-7.
    }
}

\thispagestyle{mystyle}

% \startpage{1}

% \pagestyle{plain}

\section{Introduction}

\begin{figure}[ht]
  \centering
  \includegraphics[width=\linewidth]{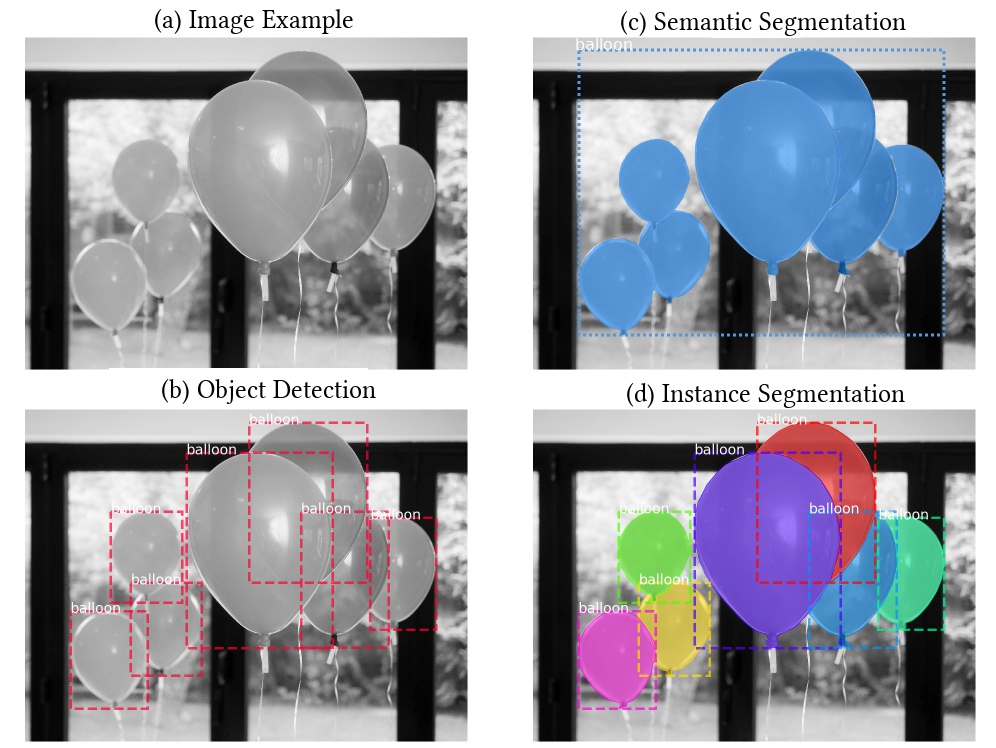}
  \caption{Difference between object detection, semantic segmentation, and instance segmentation. Adapted from~\cite{Abdulla2018website}.}
  \Description{Illustration}
  \label{fig:balloons}
\end{figure}

Object detection and semantic segmentation are common research topics in Computer Vision. Detection focus on finding and identifying different objects in an image, defining a bounding-box around them. Semantic segmentation is more challenging, aiming at not only finding objects, but also the contours at the limits of the objects. Both tasks have a wide range of applications like robotics, medical diagnosis, fashion design, mapping streets, and many others. An even more difficult task is instance semantic segmentation, where the focus is not only segmentation of classes, but individual instances of the classes. Figure~\ref{fig:balloons} demonstrates each task's objective.

Detection of objects nowadays can be made by advanced algorithms that have already surpassed human-level performance~\cite{he2015delving}. The introduction of faster detection methods, like Faster R-CNN~\cite{ren2015faster}, YOLO~\cite{redmon2016you}, and recently, Mask R-CNN~\cite{he2017mask}, made possible more accurate and real-time methods for many real-world tasks~\cite{laroca2019efficient}.

In contrast, object segmentation is still not as good as humans, as it is a much harder task to segment objects, especially in a pixel-wise fashion. One of the drawbacks of segmentation performance is the lack of available large-scale datasets and labeled images. The main cause of this scarcity lies upon the manual generation of semantic masks being laborious. For instance, for assembling the Cityscapes dataset of urban scenes semantic segmentation~\cite{cordts2016cityscapes}, an average of 90 minutes was required for labeling each image.

Thus, any effective alternative to manually labeling image pixels is relevant. Generative Adversarial Networks (GANs)~\cite{goodfellow2014generative} are possibly a viable alternative to this issue. GAN is a deep network model composed of two networks: a generator and a discriminator. The generator produces images from latent vectors, and the discriminator serves as a forgery expert, evaluating the quality and realness of the generated images. Both are trained in a zero-sum game against each other. GANs have been used in many applications for new images generation, and also segmentation masks. But how realistic are those images? 

Our main hypothesis is that a GAN can translate giraffes to birds and generate a synthetic dataset of realistic bird images. The quality assessment of bird generation in those images will be the main subject of study. Other GANs have been trained to translate sheep into giraffes, or horses into zebras, but those classes are similar to each other regarding scale and shape. A bigger challenge that we chose was to translate giraffes into birds because of the great difference between them regarding scale, shape, and texture.  The highlight of this work is that we are not turning apples into oranges, but elephants into bananas.

Qualitative and quantitative investigations were performed. The qualitative dealt with the resulting appearance of the image-to-image translation, the newly generated dataset entitled FakeSet. Five people answered a survey regarding the quality of translation, contour, and texture of the generated images. For the quantitative analysis, Mask R-CNN~\cite{he2017mask}, a state-of-the-art method for image semantic segmentation, was used for the detection and segmentation of the new bird dataset, and two common datasets of the literature. Detection and segmentation accuracy were taken into account. The goal was to show that the generated images have a similar overall performance in comparison to the real ones.

The following section presents recent works regarding image detection, segmentation, GAN, and image-to-image translation. Section 3 details the implementation, architecture, and parameters used for training the GAN and generating FakeSet. Consecutively, in section 4, the quality of FakeSet is investigated using two distinct approaches: qualitative and quantitative. Section 5 presents discussions arisen from the results obtained. Conclusions and future work proposals are compiled in the final section.

\section{Literature Review}

With the advents of superfast machines, GPUs, and lots of training data, object detection and semantic segmentation scaled to a high level of efficiency. 
One of the recent breakthroughs in object detection was provided by Faster R-CNN~\cite{ren2015faster}. This novel deep learning method introduced real-time object detection, using Region Proposal Networks (RPN). RPN is a convolutional network that detects objects' boundaries yielding probability scores and was merged with a Region-based Convolutional Network (R-CNN)~\cite{girshick2015fast} to form the final network architecture. RPN is an attention mechanism, indicating which regions R-CNN should focus (regions that might contain an object).

Mask R-CNN~\cite{he2017mask} is a state-of-the-art deep neural network for instance segmentation. This segmentation network adds a parallel path to the original Faster R-CNN, that calculates binary objects' masks, with minimal overhead. The loss function used for training aims at maximizing both detection and segmentation accuracy, providing high-quality trustworthy segmentation masks.

Mask R-CNN has been used for many general segmentation tasks, such as urban scenarios segmentation~\cite{he2017mask}, birds species categorization~\cite{wei2018} and salient object detection~\cite{krinski2019}.

To generate more examples of training, Generative Adversarial Networks (GANs) can be used. They can learn real image features and generate synthetic images from latent feature vectors (with the addition of noise). Introduced in 2014 by Goodfellow \textit{et al.} ~\cite{goodfellow2014generative}, GANs are widely used nowadays especially after advances made in the loss functions, to constrain learning and generation of more realistic images.

ALI~\cite{dumoulin2016adversarially} and BiGAN~\cite{donahue2016adversarial} proposed an inverse mapping (from image to latent vector). This intuition is very useful in image-to-image translation, when instead of generating a latent vector, an image of another domain is generated.

CoGAN ~\cite{liu2016coupled} focused in multi-domain scenarios. Without the need to pair samples from different domains, it learns image translation by training two distinct generators with weight-sharing to ensure a joint distribution.
~\cite{shrivastava2017learning}

Pix2pix~\cite{isola2017image} uses with constrained GANs to synthesize images from contours, applying styles for images, colorization. It also proposes automatic learning of mappings, using loss functions, showing the versatility of learning automatically mapping functions for image-to-image translation.

Although all the previously referenced works contributed to image-to-image translation research, one of the most groundbreaking approaches was made by CycleGAN~\cite{zhu2017unpaired}. CycleGAN introduced a constraining loss function for the generative network to maintain cycle integrity. In other words, being the target domains $A$ and $B$ and the generative networks $G_B$ and $G_A$, the following mappings are learned: $G_B: A \rightarrow B$, $G_A: B \rightarrow A$. The cycle consistency loss, restrains that any mapping from $A$ to $B$, and back to $A$ should be close to the initial region in $A$, formally: $G_A(G_B(A)) \approx A$. The same applies to $B$, $G_B(G_A(B)) \approx B$.

A recent work~\cite{li2018semantic}, used semantic-aware GANs to make images extracted from the GTA-V game more realistic, making them possible to be used for training, enhancing semantic segmentation accuracy on real street images. A great example of how data can be generated to be used as data augmentation for training a segmentation network. Ruiz et al.~\cite{ruiz2019anda} proposed a novel data augmentation technique used for segmentation that combines image inpainting and linear combinations of different images. 

Focusing in instance semantic segmentation, one of the first and most recent methods proposed is Instance-aware Generative Adversarial Network (InstaGAN)~\cite{mo2018instagan}. 
Although other GANs like CycleGAN focused in semantic segmentation, they did not take into account the individuals (instances) of the semantic classes. InstaGAN is able to preserve even instance attributes, like position in the image. Another feature is context preservation: looking direction of animals, by example, and keeping the original background. This can be achieved by jointly encoding the attributes and image information to ensure correspondence between them.

\begin{figure*}[!htpb]
  \centering
  \subfloat[The generators G translate a pair of image and its attributes (including the segmentation mask) from one domain to the other.]{
 \includegraphics[width=0.24\linewidth]{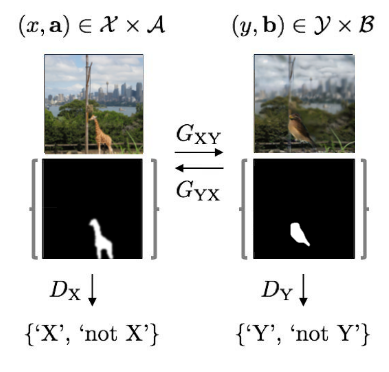}
    \label{dtdtransillust}
  }  
  \qquad
  \subfloat[A generator in action: an input image with their respective instances masks, from class $\mathcal{X}$ receives a new mask for each new instance of class $\mathcal{Y}$, based in attributes from the original image. The translated image keeps context information (background), only transforming the target instances, generating birds from a latent space of representation learned during the training stage.]{
  \includegraphics[width=0.33\linewidth]{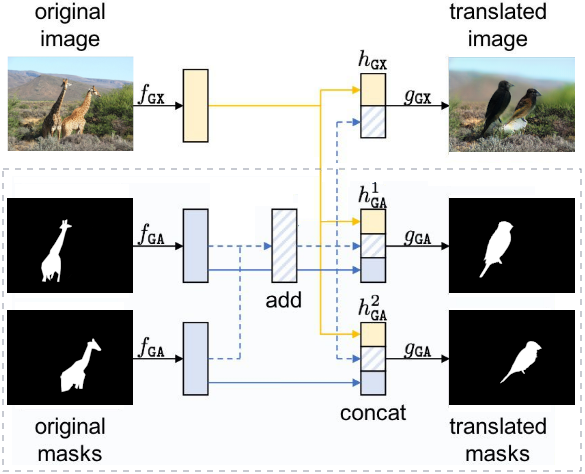}
    \label{generatorillust}
  }  
  \qquad
    \subfloat[The discriminator from class $\mathcal{Y}$ is trained to identify realistic images, and output if the image appears to be synthetic or not.]{
  \includegraphics[width=0.32\linewidth]{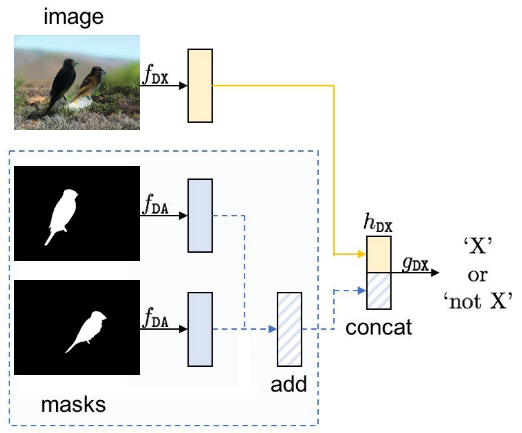}
    \label{discriminatorillust}
  }
  \caption{Illustration adapted from Mo \textit{et al.}~\cite{mo2018instagan}.}
  \label{instaganarc}
\end{figure*}

InstaGAN's generators $G_{XY}$ and $G_{YX}$ work by mapping images and attributes from one image domain $\mathcal{X}$ to the other $\mathcal{Y}$, and vice-versa. Considering a pair of image domain $\mathcal{X}$ and respective attribute domain $\mathcal{A}$, it is possible to learn a mapping to another image domain $\mathcal{Y}$ with the respective attributes related to that domain $\mathcal{B}$. Then, for a pair of image and its attributes $(x,a)  \in \mathcal{X} \times \mathcal{A}$, we will have a generator $G_{XY}$ such that: 
\begin{equation}\label{eq:generator}
    G_{XY}(x,a)= (y,b) \ \  | \ \ (y,b) \in \mathcal{Y} \times \mathcal{B}
\end{equation}
The inverse mapping is also possible, hence for a pair of image and their attributes $(y,b)  \in \mathcal{Y} \times \mathcal{B}$:
\begin{equation}\label{eq:generatorInverse}
    G_{YX}(y,b)= (x,a) \ \   | \ \  (x,a) \in \mathcal{X} \times \mathcal{A}
\end{equation}
For every image instance of the semantic domain $\mathcal{X}$, a corresponding instance of $\mathcal{Y}$ is generated, preserving attributes like position, pose and others. To translate images between domains, masks are provided as one of the attributes, so the generative network knows the corresponding attributes in the new domain, and can also preserve background (translating only the instances). At the end, each mask is translated to a new instance mask of the new domain, consequently generating segmentation masks and images containing instances of the new domain.

The discriminator receives images and masks and evaluates if the segmented instances make sense and are similar to real instances of the new domain. To accomplish that, discriminators are initially trained on real images. Figure~\ref{instaganarc} illustrates the process of instace-aware translation between domains and the discrimination process.

\section{Dataset Generation}
For the purpose of generating FakeSet, using image-to-image translation, the best option available is the use of GAN. InstaGAN~\cite{mo2018instagan} was chosen for being one of the most recent GANs and focused in semantic segmentation. Also, InstaGAN focuses in individual instances, thus preserving spatial information and context. We want to translate giraffes into birds, therefore we will need two datasets for training: one with giraffe images and the other with bird images.

\subsection{Datasets}
A dataset entitled COCO (Common objects in Context)~\cite{lin2014microsoft} was one of the two datasets used for training InstaGAN. COCO consists of 328k images of more than 91 object types. For segmentation more than 123k images are available, with 80 object categories. For our research, we only wanted one category from this dataset: giraffes. So, only giraffe images were used, resulting in 2,546 images for training and 101 for validation, all with their respective segmentation masks.

The other dataset, Caltech-UCSD Birds 200-2011~\cite{WahCUB2002011} (a.k.a., CUB-200) is focused on birds, containing 200 birds species, and more than 11k images.

Bird images were divided 80/20 for each one of 200 species of birds. In total, were used 9,414 images for training (80\%) and 2,374 for validation (20\%). No subdivisions were made according to species to enhance generalization.

\subsection{Training}

Most parameters were kept the same as the original InstaGAN paper. The images from the giraffes and birds datasets were resized to 256x256 using bilinear interpolation. The training takes almost 3 weeks on GPU (reduced to one and a half week by using 2 NVIDIA RTX 2080 GPUs). The loss functions InstaGAN uses during training include: Least-squares GAN loss~\cite{mao2017least}, cycle loss, context loss and identity loss. The different losses were weighted according to their importance in the final result. Cycle loss received a bigger weight, as it is more important than the latter to maintain equivalence between distinct domain samples. So did LSGAN, as it improves convergence, reducing possible vanishing gradient issues. More than 100 epochs were needed to achieve good visually results and stagnation of loss minimization.

\subsection{FakeSet}
\begin{figure}[!htbp]
  \centering
  \includegraphics[width=\linewidth]{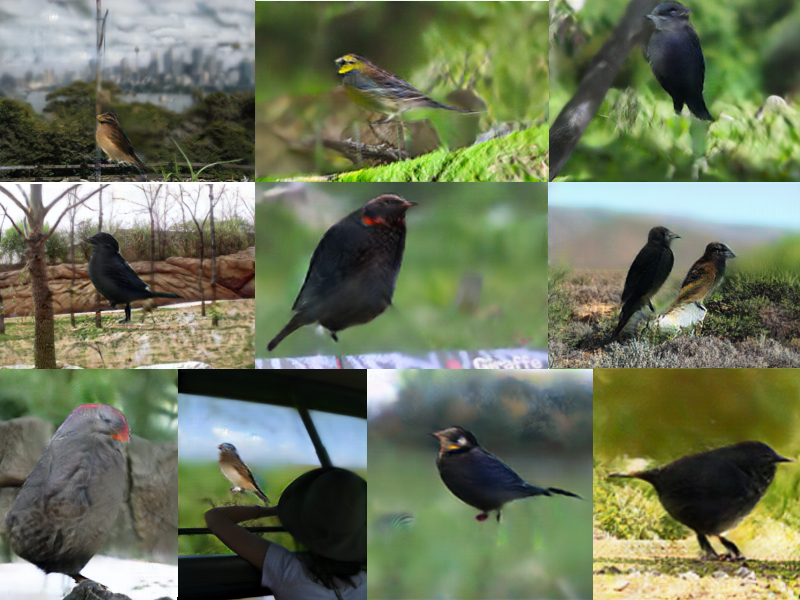}
  \caption{Generated synthetic bird images with a visually good quality (well translated birds).}
  \label{fig:goodbirds}
  \Description{Sample images with a visually good quality (well translated birds).}
\end{figure}

\begin{figure}[!htbp]
  \centering
  \includegraphics[width=\linewidth]{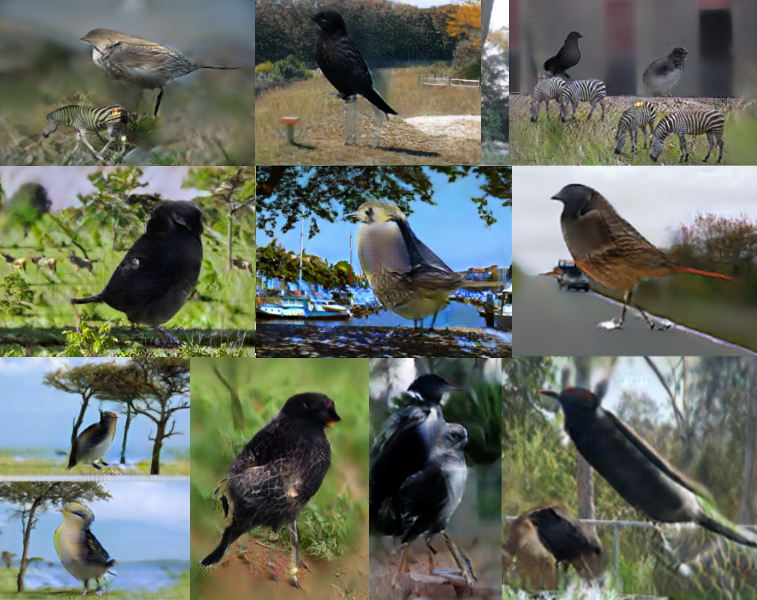}
  \caption{Generated synthetic bird images with a visually bad quality (poorly translated birds).}
   \label{fig:badbirds}
  \Description{Sample images with a visually bad quality (poorly translated birds).}
\end{figure}

The generated dataset of synthetic birds was entitled FakeSet. Some image translations performed really well, producing convincing birds and removing the original giraffes from the image, as we can see in Figure~\ref{fig:goodbirds}. However, some images did not translate well, as shown in Figure~\ref{fig:badbirds}. The translation quality of the 2,546 giraffe images to bird images will be evaluated in the next section.

\section{Dataset Evaluation}

For performance evaluation of image generation, qualitative and quantitative analyses for the generated dataset were conducted. Qualitative analysis was performed in the form of a survey. Five participants filled an online form, regarding the quality of a representative subset of FakeSet. Their answers were compiled and results were analyzed. Quantitative analysis was conducted using a pre-trained multiclass algorithm to detect and segment the new dataset images. Bird images from two common datasets were used, for results comparison. The protocols and details of both approaches will be discussed next.

\subsection{Qualitative Analysis}
In Image Processing, the main attributes of an image are: shape, contour, texture, illumination and color. In the present case, illumination and color are not really discriminative, because birds and giraffes can easily share those features. So, only shape, contour and texture were selected to be analyzed. The metrics for the qualitative analysis consisted of three main factors of quality: translation (Does the image looks like a bird or a giraffe?), contour (Were the bird limits preserved?), and texture (Does the texture look like a bird?).

\textbf{Protocol:} The Protocol for analysis consists of extracting a representative random sample from the total amount of generated images and evaluate them under the aforementioned metrics, surveying volunteers.

The minimum sample required for the population of 2,546 images can be obtained by using Cochran Equation~\cite{cochran2007sampling}:
\begin{equation}
    n_0 = \frac{Z^2pq}{e^2}
\end{equation}
In the equation, $n_0$ represents the minimum number of images required in the sampling, $Z$ represents the corresponding z-score for a desired confidence, $p$ is the estimated proportion of an attribute in the population, $q$ is $1-p$ and $e$ is the desired error.

For this experiment, we will require a confidence level of 95\% (a common value for confidence). The corresponding z-score is approximately $1.96$. The error margin will be fixed in 5\% (a common value in the literature). As the population attributes are unknown, $p$ and $q$ will both be set to achieve maximum results: $0.5$.
\begin{equation}
    n_0 = \frac{(1.96)^2 \times (0.5) \times (0.5)}{(0.05)^2 }= 385
\end{equation}
As 2,546 is a small and finite population, it is possible to reduce the sample number, using yet from Cochran:
\begin{equation}
    n = \frac{n_0}{1+\frac{n_0 -1}{N}} = \frac{385}{1+\frac{385 -1}{2546}} = 335
\end{equation}

That means, at least 335 random samples are needed to estimate the conditions of all 2,546 images generated, with 95\% confidence (and an error margin of $\pm$5\%).

\textbf{Survey:}  An intern web application was developed to inquire participants about the FakeSet's subset of 335 randomly selected images. 

Five candidates were asked to evaluate each one of the same 335 images, regarding the protocol metrics: quality of translation, contour and texture. The options were binary: ``Good'' or ``Bad'' . The final score for each attribute is calculated using the mean for all images per candidate, resulting in a score between 0 and 1 (0 being bad quality and 1 being good quality). By example, if a quality metric is equal to 0.75 for a participant, it means that 3/4 of the images performed well in that attribute and 1/4 did not. The voting results about the quality of the generated FakeSet are displayed in Table~\ref{tab:comparisonqualit}.

\begin{table}[!htbp]
  \caption{Mean quality score of 335 selected images, per participant with the respective standard deviation. Also, the estimated number of satisfactory images per attribute (and the absolute error margin) in the whole FakeSet, that contains 2,546 images.}
  \label{tab:comparisonqualit}
  \setlength{\tabcolsep}{0.8em} % for the horizontal padding
{\renewcommand{\arraystretch}{1.25}% for the vertical padding
  \begin{tabular}{lccc}
    \toprule 
      \multirow{2}{*}{\textbf{Quality Metric}} & \multicolumn{2}{c}{\textbf{Score}} &  \multirow{2}{*}{\textbf{\# Est. FakeSet ($e$=5\%)}}\\
     & Avg & Std \\
    \midrule
    \centering
    Image Translation & 0.478 & 0.175 & 1,216 ($\pm$61) \\
    \midrule
    Contour & 0.467 & 0.144 & 1,188 ($\pm$60)\\
    \midrule
     Texture & 0.651 & 0.098 & 1,657 ($\pm$83)\\
    \bottomrule
  \end{tabular}}
\end{table}

\subsection{Quantitative Analysis}

We need to verify if the images are adequate for proper detection and subsequent segmentation. We use the segmentation as finer evaluation than simple detection since it provides a pixel-wise classification. A general state-of-the-art detection and segmentation network was chosen for the task: Mask R-CNN~\cite{he2017mask}. The advantage of choosing a network trained on independent datasets from the ones used in this paper is that an impartial analysis can be performed.
Also, for comparison, two datasets that contain bird images and pre-labeled (ground-truth) segmentation masks were chosen. The evaluation protocol will be discussed next. 

\textbf{Datasets:} Two datasets were used for comparison with FakeSet: Pascal Visual Object Classes (VOC 2011)~\cite{everingham2010pascal} and Caltech-UCSD Birds 200-2011~\cite{WahCUB2002011}. VOC contains several classes, so only bird images were selected. Caltech contains more than 11k bird images, and VOC more than 250. Both datasets provide the segmentation masks, thus it is possible to investigate detection and also segmentation performance.

\textbf{Metrics:} for detection efficiency evaluation the metric used was accuracy (how many birds were detected in relation to the whole number of birds). To consider a bird detected, the confidence was set to a minimal of 80\%. To verify segmentation performance, common metrics used in the literature are: F-score and MAE (Mean Absolute Error). Another common metric is IoU. According to Gallagher \textit{et al.}~\cite{gallagher1999compah}, IoU is very similar to the F-score. For that reason, we did not measure IoU. F-score can be obtained by the following formula, with $X$ being the target class mask and $Y$ the ground-truth mask:
$$
\text{F-score} = \frac{ (1+\beta^2)\cdot|X \cap Y |}{\beta^2\cdot|X|+|Y|}
$$
We set $\beta$ to 1 as it is commonly used, obtaining:
$$
\text{F-score} = \frac{ 2|X \cap Y |}{|X|+|Y|}
$$
And the MAE can be calculated by using the following equation, where $x$ represents the pixel binary value of the segmentation mask calculated by the method, $y$ the pixel binary value of the ground-truth mask and $n$ the total number of pixels in the mask:
$$
\text{MAE} = \frac{ \sum_{i=1}^n |y_i - x_i |}{n}
$$

\textbf{Protocol:} The protocol chosen to evaluate the realness (quality) of the images consists in using a pre-trained network to evaluate detection and segmentation. 

\begin{figure}[ht]
  \centering
  \includegraphics[width=\linewidth]{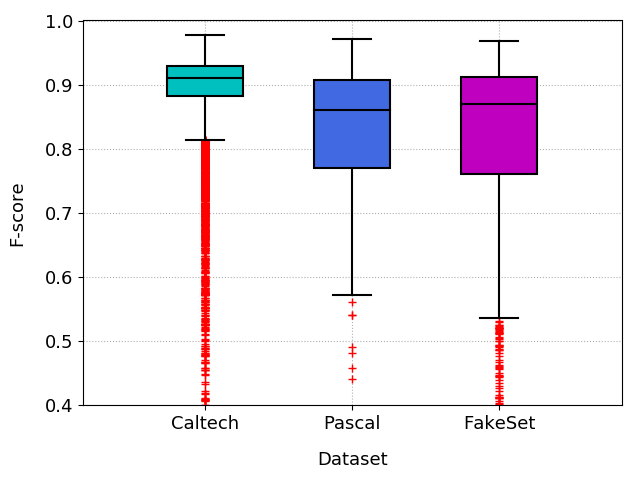}
  \caption{F-score results per dataset}
    \label{fig:f1}

  \Description{F-score results distribution using Box-plot.}
\end{figure}

FakeSet has 2,546 unique images. The evaluation demonstrated that: 289 images are trash (images that birds are not even detected), 717 are unsatisfactory (the F-score lower than 0.8) and 1,540 are satisfactory. The box plots illustrated in Figure~\ref{fig:f1} show that the distribution of F-score follow a skewed normal distribution, slanted towards best scores. Similar to the distribution of the other two datasets. These results demonstrate that most of FakeSet birds are realistic enough for being correctly detected and segmented.

Table ~\ref{tab:comparison} shows the comparison between image datasets using Mask-RCNN. Caltech refers to the Caltech-UCSD
Birds 200-2011 and Pascal to the Pascal Visual Object Classes (VOC 2011). The Mask R-CNN pre-trained with the COCO dataset performed better on the Caltech dataset since the birds are usually prominent and represent a large area of the image where in the Pascal VOC that is not always the case. 

\begin{table}
  \caption{Detection and Segmentation Accuracy.}
  \label{tab:comparison}
  \setlength{\tabcolsep}{0.5em} % for the horizontal padding
{\renewcommand{\arraystretch}{1.3}% for the vertical padding
  \begin{tabular}{lccccccc}
    \toprule 
    \multirow{2}{*}{\textbf{Metrics}} & \multicolumn{2}{c}{\textbf{Pascal}} & \multicolumn{2}{c}{\textbf{Caltech}} & \multicolumn{2}{c}{\textbf{FakeSet}}  \\
    & Avg & Std & Avg & Std & Avg & Std \\
     \midrule
     \centering
    Detection Acc. & 0.918 & -- & 0.983 & --  & 0.886 & -- \\
    \midrule
    F-score & 0.819 & 0.130 & 0.891 & 0.075 & 0.816 & 0.142 \\
    \midrule
    MAE & 0.030 & 0.035 & 0.025  & 0.024 & 0.038  & 0.036 \\

    \bottomrule
  \end{tabular}}
\end{table}

\section{Discussion}

Our qualitative findings show that most of the images are not completely realistic. However, there is a significant percentage of correctly translated images.

\begin{figure*}[!h]
  \centering

  \subfloat[Original Image]{
 \includegraphics[width=0.15\linewidth]{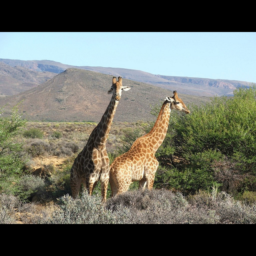}
    \label{g0Img}
  }  
  \hfill
  \subfloat[Original Mask]{
  \includegraphics[width=0.15\linewidth]{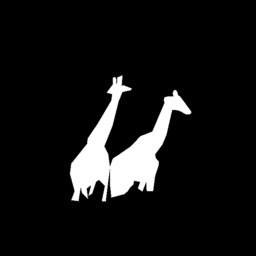}
    \label{g0Seg}
  }
  \hfill
    \subfloat[Original Image]{
 \includegraphics[width=0.15\linewidth]{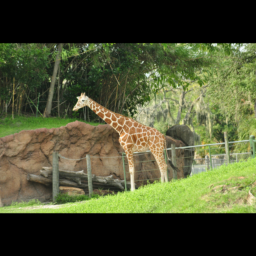}
    \label{g1Img}
  }  
  \hfill
  \subfloat[Original Mask]{
  \includegraphics[width=0.15\linewidth]{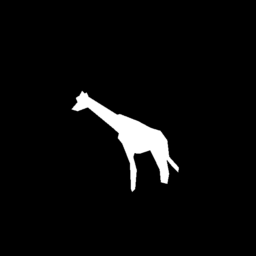}
    \label{g1Seg}
  }
\hfill
    \subfloat[Original Image]{
 \includegraphics[width=0.15\linewidth]{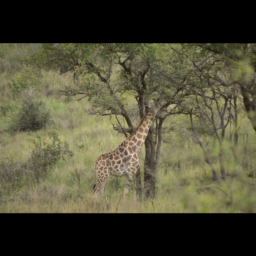}
    \label{g2Img}
  }  
  \hfill
  \subfloat[Original Mask]{
  \includegraphics[width=0.15\linewidth]{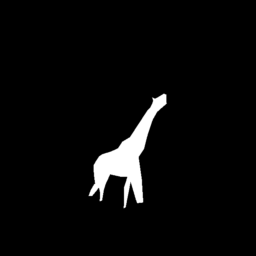}
    \label{g2Seg}
  }
  
  \subfloat[Generated Image]{
  \includegraphics[width=0.15\linewidth]{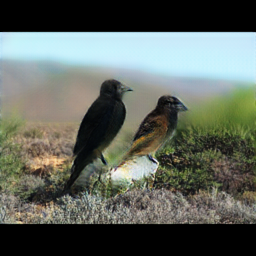}
    \label{b3Img}
  }
  \hfill
  \subfloat[Generated Mask]{
  \includegraphics[width=0.15\linewidth]{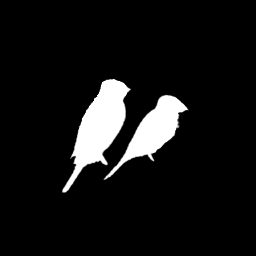}
    \label{b3Seg}
  }
  \hfill
  \subfloat[Generated Image]{
  \includegraphics[width=0.15\linewidth]{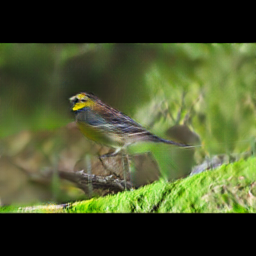}
    \label{b4Img}
  }
  \hfill
  \subfloat[Generated Mask]{
  \includegraphics[width=0.15\linewidth]{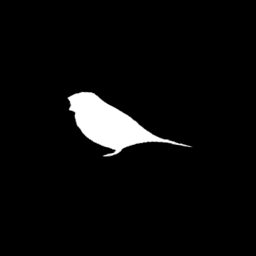}
    \label{b4Seg}
  }  
  \hfill
  \subfloat[Generated Image]{
  \includegraphics[width=0.15\linewidth]{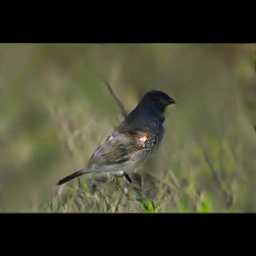}
    \label{b5Img}
  }
  \hfill
  \subfloat[Generated Mask]{
  \includegraphics[width=0.15\linewidth]{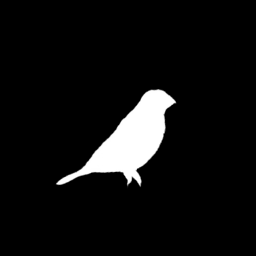}
    \label{b5Seg}
  }  
  \caption{Examples of generated translation.}
  \label{example}
\end{figure*}

In the quantitative analysis it was evident that even the images that were not utterly realistic, could still be detected and segmented with high-confidence (higher than 80\%), in a similar proportion  to other real dataset images. A correct detection and segmentation of a fake sample implies that the fake image share common features with the real images originally learned by the Deep Neural Network~(DNN), such as border, shape and texture. Figure~\ref{example} show some examples of the image-to-image translation, with the original images and respective masks, and after the translation, the resulting image, generated mask. Notice how the pose, spatial arrangement and background stays the same. We noticed that InstaGAN learned reasonably well to translate images in which giraffes have almost no occlusion and are standing sideways, producing birds on the same position. An interesting ``cheat'' that the network learned was to generate dark birds to generate contrast with the lighter backgrounds so it can fool the discriminator without having to always produce correct textures.

\section{Conclusion}
The image-to-image translation is a fairly recent research field with still many open questions, such as the proposal of new loss functions, new applications, and how to properly take advantage of the latent space representation. This field explores unsupervised learning to produce never before seen samples without the direct guidance of an expert. In the present work we have shown that bird images can be generated from giraffe images and produce labeled data using GAN. As seen in the evaluation, a portion of them will be realistic, and most of them will be able to ``fool'' in a loose sense since Mask R-CNN was not designed to detected forgery, a state-of-the-art network as being a real sample.

Evaluation of the synthetic images as a reliable source of labeled data for supervised training was left for future work. Also, multi-domain translation can be explored, enabling the generation of multiple domains from a single image, such as giraffe to elephant, to bird, to sheep, and vice-versa by a single model.

\begin{acks}
The authors would like to thank the Coordination for the Improvement of Higher Education Personnel (CAPES) for the Masters scholarship. We gratefully acknowledge the founders of the publicly available datasets and the support of NVIDIA Corporation, donating the GPUs used for this research.
\end{acks}

% \bibliographystyle{ACM-Reference-Format}
\bibliographystyle{unsrtnat}
\bibliography{sample-base}

\begin{thebibliography}{26}
\providecommand{\natexlab}[1]{#1}
\providecommand{\url}[1]{\texttt{#1}}
\expandafter\ifx\csname urlstyle\endcsname\relax
  \providecommand{\doi}[1]{doi: #1}\else
  \providecommand{\doi}{doi: \begingroup \urlstyle{rm}\Url}\fi

\bibitem[Abdulla(2018)]{Abdulla2018website}
Waleed Abdulla.
\newblock Splash of color: Instance segmentation with {Mask R-CNN} and
  tensorflow, 2018.
\newblock URL
  \url{https://engineering.matterport.com/splash-of-color-instance-segmentation-with-mask-r-cnn-and-tensorflow-7c761e238b46}.

\bibitem[He et~al.(2015)He, Zhang, Ren, and Sun]{he2015delving}
Kaiming He, Xiangyu Zhang, Shaoqing Ren, and Jian Sun.
\newblock Delving deep into rectifiers: Surpassing human-level performance on
  imagenet classification.
\newblock In \emph{Proceedings of the IEEE international conference on computer
  vision}, pages 1026--1034, 2015.

\bibitem[Ren et~al.(2015)Ren, He, Girshick, and Sun]{ren2015faster}
Shaoqing Ren, Kaiming He, Ross Girshick, and Jian Sun.
\newblock Faster {R-CNN}: Towards real-time object detection with region
  proposal networks.
\newblock In \emph{Advances in neural information processing systems}, pages
  91--99, 2015.

\bibitem[Redmon et~al.(2016)Redmon, Divvala, Girshick, and
  Farhadi]{redmon2016you}
Joseph Redmon, Santosh Divvala, Ross Girshick, and Ali Farhadi.
\newblock You only look once: Unified, real-time object detection.
\newblock In \emph{Proceedings of the IEEE conference on computer vision and
  pattern recognition}, pages 779--788, 2016.

\bibitem[He et~al.(2017)He, Gkioxari, Doll{\'a}r, and Girshick]{he2017mask}
Kaiming He, Georgia Gkioxari, Piotr Doll{\'a}r, and Ross Girshick.
\newblock Mask {R-CNN}.
\newblock In \emph{Proceedings of the IEEE international conference on computer
  vision}, pages 2961--2969, 2017.

\bibitem[{Laroca} et~al.(2019){Laroca}, {Zanlorensi}, {Gon{\c{c}}alves},
  {Todt}, {Schwartz}, and {Menotti}]{laroca2019efficient}
R.~{Laroca}, L.~A. {Zanlorensi}, G.~R. {Gon{\c{c}}alves}, E.~{Todt}, W.~R.
  {Schwartz}, and D.~{Menotti}.
\newblock An efficient and layout-independent automatic license plate
  recognition system based on the {YOLO} detector.
\newblock \emph{arXiv preprint}, arXiv:1909.01754:\penalty0 1--14, 2019.

\bibitem[Cordts et~al.(2016)Cordts, Omran, Ramos, Rehfeld, Enzweiler, Benenson,
  Franke, Roth, and Schiele]{cordts2016cityscapes}
Marius Cordts, Mohamed Omran, Sebastian Ramos, Timo Rehfeld, Markus Enzweiler,
  Rodrigo Benenson, Uwe Franke, Stefan Roth, and Bernt Schiele.
\newblock The cityscapes dataset for semantic urban scene understanding.
\newblock In \emph{Proceedings of the IEEE conference on computer vision and
  pattern recognition}, pages 3213--3223, 2016.

\bibitem[Goodfellow et~al.(2014)Goodfellow, Pouget-Abadie, Mirza, Xu,
  Warde-Farley, Ozair, Courville, and Bengio]{goodfellow2014generative}
Ian Goodfellow, Jean Pouget-Abadie, Mehdi Mirza, Bing Xu, David Warde-Farley,
  Sherjil Ozair, Aaron Courville, and Yoshua Bengio.
\newblock Generative adversarial nets.
\newblock In \emph{Advances in neural information processing systems}, pages
  2672--2680, 2014.

\bibitem[Girshick(2015)]{girshick2015fast}
Ross Girshick.
\newblock Fast {R-CNN}.
\newblock In \emph{Proceedings of the IEEE international conference on computer
  vision}, pages 1440--1448, 2015.

\bibitem[Wei et~al.(2018)Wei, Xie, Wu, and Shen]{wei2018}
Xiu-Shen Wei, Chen-Wei Xie, Jianxin Wu, and Chunhua Shen.
\newblock {Mask-CNN}: Localizing parts and selecting descriptors for
  fine-grained bird species categorization.
\newblock \emph{Pattern Recognition}, 76:\penalty0 704 -- 714, 2018.
\newblock ISSN 0031-3203.
\newblock \doi{https://doi.org/10.1016/j.patcog.2017.10.002}.
\newblock URL
  \url{http://www.sciencedirect.com/science/article/pii/S0031320317303990}.

\bibitem[Krinski et~al.(2019)Krinski, Ruiz, Machado, and Todt]{krinski2019}
Bruno~A. Krinski, Daniel~V. Ruiz, Guilherme~Z. Machado, and Eduardo Todt.
\newblock Masking salient object detection, a mask region-based convolutional
  neural network analysis for segmentation of salient objects.
\newblock In \emph{2019 Latin American Robotic Symposium, 2019 Brazilian
  Symposium on Robotics (SBR) and 2019 Workshop on Robotics in Education
  (WRE)}, October 2019.

\bibitem[Dumoulin et~al.(2016)Dumoulin, Belghazi, Poole, Mastropietro, Lamb,
  Arjovsky, and Courville]{dumoulin2016adversarially}
Vincent Dumoulin, Ishmael Belghazi, Ben Poole, Olivier Mastropietro, Alex Lamb,
  Martin Arjovsky, and Aaron Courville.
\newblock Adversarially learned inference.
\newblock \emph{arXiv preprint arXiv:1606.00704}, 2016.

\bibitem[Donahue et~al.(2016)Donahue, Kr{\"a}henb{\"u}hl, and
  Darrell]{donahue2016adversarial}
Jeff Donahue, Philipp Kr{\"a}henb{\"u}hl, and Trevor Darrell.
\newblock Adversarial feature learning.
\newblock \emph{arXiv preprint arXiv:1605.09782}, 2016.

\bibitem[Liu and Tuzel(2016)]{liu2016coupled}
Ming-Yu Liu and Oncel Tuzel.
\newblock Coupled generative adversarial networks.
\newblock In \emph{Advances in neural information processing systems}, pages
  469--477, 2016.

\bibitem[Shrivastava et~al.(2017)Shrivastava, Pfister, Tuzel, Susskind, Wang,
  and Webb]{shrivastava2017learning}
Ashish Shrivastava, Tomas Pfister, Oncel Tuzel, Joshua Susskind, Wenda Wang,
  and Russell Webb.
\newblock Learning from simulated and unsupervised images through adversarial
  training.
\newblock In \emph{Proceedings of the IEEE conference on computer vision and
  pattern recognition}, pages 2107--2116, 2017.

\bibitem[Isola et~al.(2017)Isola, Zhu, Zhou, and Efros]{isola2017image}
Phillip Isola, Jun-Yan Zhu, Tinghui Zhou, and Alexei~A Efros.
\newblock Image-to-image translation with conditional adversarial networks.
\newblock In \emph{Proceedings of the IEEE conference on computer vision and
  pattern recognition}, pages 1125--1134, 2017.

\bibitem[Zhu et~al.(2017)Zhu, Park, Isola, and Efros]{zhu2017unpaired}
Jun-Yan Zhu, Taesung Park, Phillip Isola, and Alexei~A Efros.
\newblock Unpaired image-to-image translation using cycle-consistent
  adversarial networks.
\newblock In \emph{Proceedings of the IEEE international conference on computer
  vision}, pages 2223--2232, 2017.

\bibitem[Li et~al.(2018)Li, Liang, Jia, and Xing]{li2018semantic}
Peilun Li, Xiaodan Liang, Daoyuan Jia, and Eric~P Xing.
\newblock {Semantic-aware Grad-GAN for Virtual-to-Real Urban Scene Adaption}.
\newblock \emph{arXiv preprint arXiv:1801.01726}, 2018.

\bibitem[Ruiz et~al.(2019)Ruiz, Krinski, and Todt]{ruiz2019anda}
Daniel~V. Ruiz, Bruno~A. Krinski, and Eduardo Todt.
\newblock {ANDA}: A novel data augmentation technique applied to salient object
  detection.
\newblock In \emph{2019 19th International Conference on Advanced
  Robotics~(ICAR)}, December 2019.

\bibitem[Mo et~al.(2019)Mo, Cho, and Shin]{mo2018instagan}
Sangwoo Mo, Minsu Cho, and Jinwoo Shin.
\newblock {InstaGAN:} {Instance}-aware {Image-to-Image} {Translation}.
\newblock In \emph{International Conference on Learning Representations}, 2019.
\newblock URL \url{https://openreview.net/forum?id=ryxwJhC9YX}.

\bibitem[Lin et~al.(2014)Lin, Maire, Belongie, Hays, Perona, Ramanan,
  Doll{\'a}r, and Zitnick]{lin2014microsoft}
Tsung-Yi Lin, Michael Maire, Serge Belongie, James Hays, Pietro Perona, Deva
  Ramanan, Piotr Doll{\'a}r, and C~Lawrence Zitnick.
\newblock {Microsoft COCO: Common Objects in Context}.
\newblock In \emph{European conference on computer vision}, pages 740--755.
  Springer, 2014.

\bibitem[Wah et~al.(2011)Wah, Branson, Welinder, Perona, and
  Belongie]{WahCUB2002011}
C.~Wah, S.~Branson, P.~Welinder, P.~Perona, and S.~Belongie.
\newblock {The Caltech-UCSD Birds-200-2011 Dataset}.
\newblock Technical Report CNS-TR-2011-001, California Institute of Technology,
  2011.

\bibitem[Mao et~al.(2017)Mao, Li, Xie, Lau, Wang, and
  Paul~Smolley]{mao2017least}
Xudong Mao, Qing Li, Haoran Xie, Raymond~YK Lau, Zhen Wang, and Stephen
  Paul~Smolley.
\newblock Least squares generative adversarial networks.
\newblock In \emph{Proceedings of the IEEE International Conference on Computer
  Vision}, pages 2794--2802, 2017.

\bibitem[Cochran(2007)]{cochran2007sampling}
William~G Cochran.
\newblock \emph{Sampling techniques}.
\newblock John Wiley \& Sons, 2007.

\bibitem[Everingham et~al.(2010)Everingham, Van~Gool, Williams, Winn, and
  Zisserman]{everingham2010pascal}
Mark Everingham, Luc Van~Gool, Christopher~KI Williams, John Winn, and Andrew
  Zisserman.
\newblock The {PASCAL Visual Object Classes (VOC)} challenge.
\newblock \emph{International journal of computer vision}, 88\penalty0
  (2):\penalty0 303--338, 2010.

\bibitem[Gallagher(1999)]{gallagher1999compah}
E~Gallagher.
\newblock Compah documentation.
\newblock \emph{User's Guide and application}, 1999.
\newblock URL \url{http://www.es.umb.edu/edgwebp.htm}.

\end{thebibliography}


\end{document}